# Energy Management of Multi-mode Hybrid Electric Vehicles based on Hand-shaking Multi-agent Learning

Min Hua[1], Cetengfei Zhang[1], Zhi Li[2], Xiaoli Yu[2], Hongming Xu[1], Quan Zhou[1,2*]

1 School of Engineering, University of Birmingham, Birmingham, B15 2TT, UK

2 State Key Laboratory of Clean Energy Utilization, Zhejiang University, Hangzhou 310027, China

**ABSTRACT**

The future transportation system will be a multi-agent network where connected AI agents can work together to address the grand challenges in our age, e.g., mitigation of real-world driving energy consumption. Distinguished from the existing research on vehicle energy management, which decoupled multiple inputs and multiple outputs (MIMO) control into single-output (MISO) control, this paper studied a multi-agent deep reinforcement learning (MADRL) framework to deal with multiple control outputs simultaneously. A new hand-shaking strategy is proposed for the DRL agents by introducing an independence ratio, and a parametric study is conducted to obtain the best setting for the MADRL framework. The study suggested that the MADRL with an independence ratio of 0.2 is the best, and more than 2.4% of energy can be saved over the conventional DRL framework.

## 1. INTRODUCTION

To meet the global requirements in decarbonization, it is necessary to develop electrified vehicles including plug-in hybrids, battery electric, and fuel cell vehicles [1], among which, plug-in hybrid electric vehicles (PHEVs) are the lowest cost solution to alleviate vehicle emission concerns, such as $NO_x$, and $CO_2$ emissions [2]. The energy management system (EMS) is a critical function module for electrified vehicles, and it should be dedicated to various powertrain architectures and thus is capable to maximize energy efficiency while maintaining the health of the powertrain components [3][4].

From the existing literature, there are three main categories of control strategies for the EMS, i.e., the rule-based methods, the optimization-based methods, and the learning-based methods [5–7]. The rule-based strategies normally implement deterministic rules or fuzzy logic that are founded on the parameters of the vehicle and expert knowledge to control the powertrain. The rule-based control is computationally efficient and easy to be applied in real-time control [8].

The optimization-based strategies include the model predictive control (MPC) [9], dynamic programming (DP) [10], pontryagin's minimum principle (PMP)[11], and equivalent consumption minimization strategy (ECMS) [12]. They provide access to optimize vehicle performance in certain conditions. The main drawback of the optimization-based strategies is that they have limited adaptability to real-world conditions especially for dramatically changing conditions, and they are tough to obtain good results for multi-objective and multi-mode optimization problems since they require heavy-duty computation to resolve the control models [13].

The learning-based methods are emerging in recent years to enable the optimization of control policies in real-world operations [14]. Deep Reinforcement Learning (DRL), a combination of deep learning (DL) and reinforcement learning (RL), has been shown more flexible and generally applicable in the control of HEVs and PHEVs [15]. The recent development of HEVs and PHEVs involves more diverse architectures and an increasing number of powertrain components. This requires dedicated approaches to design the EMS for different types of PHEVs [16].

There is a lot of work developing RL-based EMS based on conventional RL algorithms. Considering three aspects including the energy consumption, the real-time performance, and the adaptability in different scenarios, an RL-based online EMS is presented based on the power transition probability matrices updated by the new driving cycle and Kullback-Leibler (KL) divergence rate [17]. A deep Q-network (DQN)-based EMS combining two distributed DRL, including asynchronous advantage actor-critic (A3C) and distributed proximal policy optimization (DPPO) for emissions control, is proposed to achieve near-optimal fuel economy with excellent computational efficiency [18]. A hierarchical power splitting strategy, which incorporates reinforcement learning and ECMS with an adaptive fuzzy filter, is proposed to reduce the state-action space for the EMS of fuel cell hybrid electric vehicles [19]. DRL-based control strategies combined with other advanced algorithms have also been employed in the EMS. They provided much better control rules compared to the tabular Q-learning method and rule-based scheme, and their dependency on the reduced-order models, which are critical to ECMS and MPC, can be reduced due to their model-free feature [20].

Multi-mode PHEV is a new powertrain topology developed in recent years to allow the vehicle operates in pure battery mode, series hybrid mode, and parallel hybrid mode flexibility for maximum fuel economy [21]. This powertrain topology has been widely adopted by OEMs and T1 suppliers worldwide, e.g., Honda, BYD, and MAHLE [22]. Distinguished from the control of series or parallel HEVs, the control of the engine, generator, and traction motor cannot be coupled, and multi-inputs-multi-outputs (MIMO) control is required. This brings new challenges for the development of RL-based EMS since the conventional single-agent RL algorithms are not well suited for the MIMO control [23].

The requirements for developing advanced RL-based EMS for the multi-mode PHEV motivate the evolution from single-agent learning to multi-agent learning because multi-agent DRL (MADRL) offers a feasible path to solve the MIMO control problems [24]. MADRL emphasizes the behaviors of multiple learning agents coexisting in a common environment with different collaboration modes, and there are three working modes between the RL agents: 1) cooperative mode, 2) competitive mode, and 3) a mixture of the two [25]. In cooperative scenarios, agents work together to maximize a shared long-term return; in contrast, in competitive scenarios, agents' returns typically add up to zero; in mixed scenarios, there are general sum returns in both cooperative and competitive agents [26]. Because the RL-based EMS for multi-mode PHEV has yet to be proposed, the work is done with two new contributions: 1) a hand-shaking strategy is proposed for the DRL agents by introducing an independence ratio, and 2) a parametric study is conducted to obtain the best setting for the MADRL framework.

The rest of this paper is organized as follows: Section 2 formulates the control of a multi-mode PHEV as a MIMO control problem through mathematical modelling. Then the structure of hand-shaking multi-agent learning is proposed in Section 3 with DDPG-based EMS introduced as the baseline method. Section 4 conducts simulation experiments and analyzes the results for validation and evaluation, followed by the conclusions in Section 5.

## 2. MIMO CONTROL IN A MULTI-MODE PHEV

The architecture of the multi-mode PHEV studied in this paper is shown in Fig. 1. The motor generator (MG1) and the engine work together to keep the battery's SoC constant for longer driving distances, and the other motor (MG2) and engine are the power sources to drive and brake the vehicle. The multi-mode PHEV has different working modes through different mechanical connections like clutch, including series hybrid mode, parallel hybrid mode, series-parallel hybrid mode, and regenerative brake as shown in Fig. 1 with different color lines to achieve the charge-sustaining, which makes the EMS more complicated.

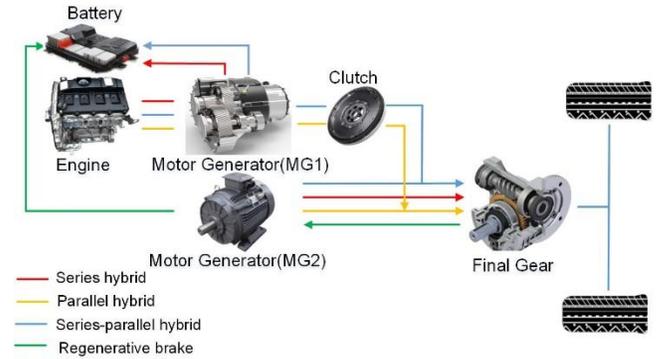

Fig. 1 Configuration of the multi-mode HEV powertrain

### 2.1 The energy flow model

In general, the energy flow of the vehicle is modeled based on longitudinal vehicle dynamics, and the force demand, $F_{dem}(t)$, the power demand, $P_{dem}(t)$, and the torque demand, $T_{dem}(t)$ can be calculated by

$$F_{dem}(t) = mgf + \frac{1}{2}\rho A_f C_d v^2(t) + mg\alpha + ma \quad (1)$$
$$P_{dem}(t) = F_{dem}(t) \cdot v \quad (2)$$
$$T_{dem}(t) = F_{dem}(t) \cdot R \quad (3)$$

where $m$ is the vehicle mass; $a$ is the vehicle acceleration, $g$ is the gravity acceleration; $f$ is the rolling resistance coefficient; $\rho$ is the air density; $A_f$ is the front area of the vehicle; $C_d$ is the air resistance coefficient; $v$ is the longitudinal velocity; $\alpha$ is the road slope; and $R$ is the wheel radius.

In the series mode, the clutch is disengaged, the MG2 solely provides the powertrain for the vehicle to achieve the power demand, and the engine and MG1 works together to keep the SOC, thus, the energy flow is described as:

$$\left.\begin{array}{l}P_{dem}(t) = P_{mot2}(t) \\ T_{dem}(t) = i_2 \cdot T_{mot2}(t) \\ P_{eng}(t) = \dfrac{n_{eng}(t) \cdot T_{eng}(t)}{9550} \\ P_{eng}(t) = P_{mot1}(t) \\ n_{eng}(t) = n_{mot1}(t)\end{array}\right\} \quad (4)$$

In the parallel mode, the engine and MG2 both drive the powertrain by the clutch engagement to achieve the power demand, thus, the energy flow is described as:

$$\left.\begin{array}{l} P_{dem}(t) = P_{mot2}(t) + (P_{eng}(t) - P_{mot1}(t)) \\ T_{dem}(t) = i_1 \cdot \left(T_{eng}(t) - T_{mot1}(t)\right) + i_2 \cdot T_{mot2}(t) \\ P_{eng}(t) = \dfrac{n_{eng}(t) \cdot T_{eng}(t)}{9550} \\ n_{eng}(t) = n_{mot1}(t) \end{array}\right\} \quad (5)$$

where, $P_{mot1,2}(t)$ and $P_{eng}(t)$ are the electric power of MG1 and MG2, respectively; $n_{mot1}(t)$, $n_{mot2}(t)$, and $n_{eng}(t)$ are the rotational speed of MG1, MG2, and the engine, respectively; the MG1 and MG2 torque are separately $T_{mot1,2}(t)$ and $T_{eng}(t)$ is the engine torque; $i_1$ is the transmission ratio of the gearbox to MG1, $i_2$ is the final ratio of the gearbox from MG2 to the wheels.

The energy of multi-mode PHEV is from the battery and the engine (fuel tank), and the total power loss mainly consists of the engine loss, $Loss_{eng}(t)$, and battery loss, $Loss_{batt}(t)$, which can be calculated by

$$\left.\begin{array}{l} P_{loss}(t) = Loss_{eng}(t) + Loss_{batt}(t) \\ Loss_{eng}(t) = \dot{m}_f(t) \cdot H_f - \dfrac{n_{eng}(t) \cdot T_{eng}(t)}{9550} \\ Loss_{batt}(t) = R \cdot I_{batt}(t)^2 \end{array}\right\} \quad (6)$$

where, $P_{loss}$ is the total power loss; $H_f$ is the heat value of fuel ($H_f = 43.5 \times 10^6$ J/kg); and $R$ is the equivalent internal resistance in the battery model.

*1) Engine model*

The engine model is used to determine the fuel consumption rate $\dot{m}_f$ (g/s) based on a 2D look-up table, which is a function of the engine speed, $n_{eng}(t)$, and the engine torque, $T_{eng}(t)$, by

$$\dot{m}_f(t) = f(n_{eng}(t), T_{eng}(t)) \quad (7)$$

*2) Motor models*

The power demands of MG1 and MG2 are modelled based on two quasi-static energy efficiency maps $\eta_1$ and $\eta_2$, respectively, as follows

$$P_{mot1}(t) = \dfrac{n_{mot1}(t) \cdot T_{mot1}(t)}{9550} \cdot \eta_1(n_{mot1}(t), T_{mot1}(t)) \quad (8)$$

$$P_{mot2}(t) = \begin{cases} \dfrac{n_{mot2}(t) \cdot T_{mot2}(t)}{9550} \cdot \eta_2(n_{mot2}(t), T_{mot2}(t)); & T_{mot2}(t) > 0 \\ \dfrac{n_{mot2}(t) \cdot T_{mot2}(t)}{9550} \cdot \dfrac{1}{\eta_2(n_{mot2}(t), T_{mot2}(t))}; & T_{mot2}(t) \leq 0 \end{cases} \quad (9)$$

*3) Battery model*

The battery model is established by an equivalent circuit model as follows

$$\left.\begin{array}{l} P_{batt}(t) = P_{mot2}(t) - P_{mot1}(t) \\ P_{batt}(t) = U \cdot I_{batt}(t) - R \cdot I_{batt}^2(t) \\ I_{batt}(t) = \dfrac{U - \sqrt{U^2 - 4 \cdot R \cdot P_{batt}(t)}}{2 \cdot R} \\ SoC(t) = SoC(0) - \dfrac{\int_0^t I_{batt}(t)dt}{Q_{batt}} \end{array}\right\} \quad (10)$$

where, $P_{batt}(t)$ is the output power in the charge-discharge process, $SoC(0)$ is the initial SoC value, $I_{batt}(t)$ is the current of the battery, and $Q_{batt}$ is the nominal battery capacity ($Q_{batt}$ =54.3 Ah). Since this research mainly focuses on charge-sustaining control, the battery temperature and aging dynamics are ignored. The open-circuit voltage (OCV) *U* (*U* = 350V) and the battery internal resistance *R* (*R* = 0.15Ω) are both constant.

### 2.2 The multiple inputs and multiple-output (MIMO) energy management controller

As illustrated in Fig.2. the energy flow of the studied vehicle is managed by a multiple inputs and multiple-output (MIMO) controller. By observing the battery Soc and the overall torque demand as the control inputs, the MIMO controller calculates the torque demands for MG1, MG2, and the engine, respectively, to general sufficient torque to drive the vehicle while maintaining the battery SoC.

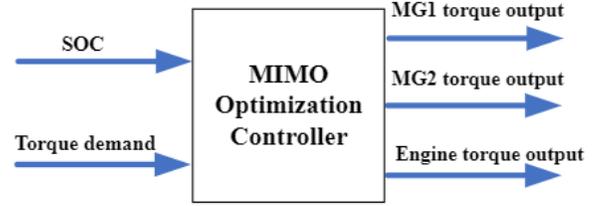

Fig. 2 MIMO control architecture

The work process of the MIMO controller is to resolve an optimization problem defined as follows：

$$Minimize \ P_{loss}(\boldsymbol{u}_{mot1}, \boldsymbol{u}_{mot2}, T_{dem}) \ and \ \Delta SOC(\boldsymbol{u}_{mot1}, \boldsymbol{u}_{mot2}, T_{dem})$$

$$s.t. \begin{cases} Loss_{eng}(t) = \dot{m}_f(t) \cdot H_f - \dfrac{n_{eng}(t) \cdot T_{eng}(t)}{9550} \\ Loss_{batt}(t) = R \cdot I_{batt}(t)^2 \\ SoC(t) = SoC(0) - \dfrac{\int_0^t I_{batt}(t)dt}{Q_{batt}} \\ SoC^- \leq SoC(t) \leq SoC^+ \\ and \ other \ physical \ constraints \end{cases} \quad (11)$$

where, the overall power loss, $P_{loss}$, and the SoC difference, $\Delta SOC$, are two objectives which need to be minimized; the MG1 torque command, $u_{mot1}(t)$, and the MG2 torque command, $u_{mot2}(t)$, are the optimization variables to be determined during the real-time control; and the optimization should be subjected to the vehicle energy flow models and other physical constraints of the powertrain system and subsystems.

## 3. HAND-SHAKING MULTI-AGENT LEARNING

To resolve the optimization problem defined in Eq. 11, this paper proposes a hand-shaking multi-agent learning scheme as shown in Fig.3 which involves two DDPG agents minimizing the fuel consumption and battery usage through the torque control of MG1 and MG2 simultaneously. Generally, the learning process starts with the observation of state variables. Then, the agent uses an actor network to generate a control action and implements it on the vehicle system to measure a reward value. By recording the state variables, action variables, and reward variables, a critic network will be trained to update the actor network. The main differences between the multi-agent learning system and the single-agent learning system (baseline) are summarized in Table I. Details of the main components of both learning systems are described as follows.

TABLE I
COMPARISON OF SINGLE-AGENT AND MULTI-AGENT SYSTEMS

| | Single-agent | Multi-agent |
|---|---|---|
| No. of Agents | 1 | 2 |
| States | $T_{dem}(t), SoC(t)$ | $T_{dem}(t), SoC(t)$ |
| Action(s) | $u_{mot1}(t)$ | $u_{mot1}(t), u_{mot2}(t)$ |
| Reward | Weighted sum | Two unique functions with different preference |

### 3.1 Agent(s)

The main difference between the single-agent learning system and the multi-agent system is the number of learning agents, i.e., the single-agent system only has one learning agent while the multi-agent system has more than two agents. The learning agent is a multi-input and single-output (MISO) control model that has the capability of self-learning for the development of control policy. It can be developed based on Q-learning, deep Q-learning, or other reinforcement learning algorithms. In this study, the environment states and action variables are continuously varying, therefore, the agent is developed based on the Deep Deterministic Policy Gradient (DDPG) algorithm. The proposed multi-agent system includes two agents with different reward preferences for the two optimization objectives, and each of the agents generates the control signal for MG1 and MG2, respectively.

### 3.2 States and actions

In this study, both single-agent system and multi-agent system monitor vehicle torque demands and battery SoC values as the state variables in a two-dimensional vector space as

$$s(t) = [T_{dem}(t), SoC(t)] \quad (12)$$

where $s(t)$ is the current state at the $t^{th}$ time step; $T_{dem}(t)$ is the power demand value at the $t^{th}$ step.

Since the single agent system can only output a single control action, $a_s(t)$, this study uses the DDPG algorithm in the single system to compute the control command $u_{mot1}(t)$ for MG1,

$$a_s(t) = u_{mot1}(t) \quad (13)$$

And the control commands $u_{mot1}(t)$ and $u_{mot2}(t)$ of MG1 and MG2, and the control command $u_{eng}(t)$ of the engine can be calculated by

$$\begin{aligned} T_{eng} &= u_{mot1}(t) \cdot T_{mot1\_max} + T_{GB} \\ T_{dem} &= u_{mot2}(t) \cdot T_{mot2\_max} + T_{eng} \\ u_{eng}(t) &= \frac{u_{mot1}(t) \cdot T_{mot1\_max} + T_{GB}}{T_{eng\_max}} \end{aligned} \quad (14)$$

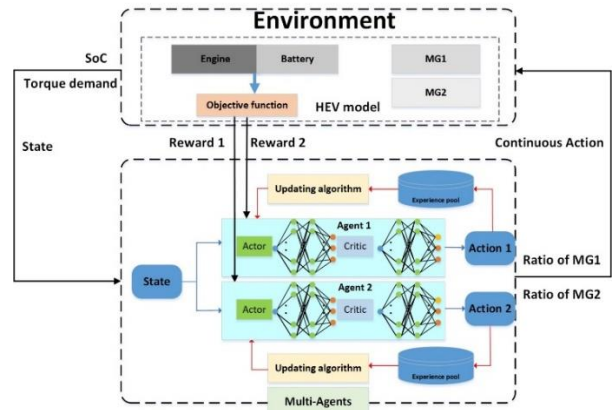

Fig.3. DDPG-based EMS with multi-agent learning

where, $T_{mot1\_max}$ is the maximum torque of the MG1; $T_{eng\_max}$ is the maximum torque that can be supplied by the engine; and $T_{dem}$ is the torque demand for driving and braking the vehicle; $T_{GB}$ is the torque of the output shaft provided by the engine when MG2 output torque cannot meet the requirement of the total torque output.

For the proposed multi-agent system that has two DDPG agents, the actions as the output of the multi-agent system can be

$$a_m(t) = [u_{mot1}(t), u_{mot2}(t)] \quad (15)$$

where $u_{mot1}(t)$ is the output of the first DDPG agent for control of MG1 while $u_{mot2}(t)$ is the output of the second DDPG agent for control of MG2. And the engine control command can be calculated using Eq.14 as well.

Both $a_s(t)$ and $a_m(t)$ are calculated following a rolling process of exploration and exploitation as defined in the DDPG algorithm [18].

*3.3 Reward functions and hand-shaking design*

The single-agent system implements a weighted sum method to incorporate the optimization objectives by

$$r_s(t) = -\alpha P_{loss}(t) - \beta |SoC_{ref} - SoC(t)| \quad (16)$$

where $\alpha$ is a scaling factor; $SoC_{ref}$ is the target battery SoC value to be maintained during the driving; and $\beta$ is a conditional weight factor, which yields

$$\beta = \begin{cases} 0, & SoC(t) \geq SoC_{ref} \\ 2, & SoC(t) < SoC_{ref} \end{cases} \quad (17)$$

The conditional weight factor will allow the DDPG agent has higher priority in minimizing fuel consumption when the SoC level is high.

The proposed multi-agent learning system provides access to comprehensive evaluations of the optimization objectives by incorporating global reward ($r_{global}$), and local rewards ($r_{local,1}$ and $r_{local,2}$) in a hand-shaking manner through the independence ratio $R_{ind}$ as

$$\begin{cases} r_{m1} = R_{ind} * r_{global} + r_{local,1} \\ r_{m2} = R_{ind} * r_{global} + r_{local,2} \end{cases} \quad (18)$$

where $r_{m1}$ and $r_{m2}$ are the rewards for the first DDPG agent and the second DDPG agent, respectively. Since, minimizing the power loss is the main optimization objective, the power loss value is the element for the global reward function,

$$r_{global} = -P_{loss}(t) \quad (19)$$

Two local reward functions are designed to balance the usages of the ICE engine and the battery with two DDPG agents. $r_{local,1}$ and $r_{local,2}$ are allocated for the first DDPG agent and the second DDPG agent respectively, and they can be calculated by:

$$\begin{cases} r_{local,1} = -\beta |SoC_{ref} - SoC(t)| \\ r_{local,2} = -\alpha Loss_{eng}(t) \end{cases} \quad (20)$$

where $\alpha$ is the scaling factor and $\beta$ is the weighting factor. In this research, $\alpha$ and $\beta$ in the multi-agent system are set as same values with the single-agent system.

## 4. RESULTS AND DISCUSSION

Based on a software-in-the-loop testing platform built in MATLAB/Simulink, this section presents the learning performance of the multi-agent system with different handshaking modes defined by different $R_{ind}$ values. This section will also compare the performances of two PHEVs that are controlled by the single-agent system and the proposed multi-agent system. Both the single-agent system and the multi-agent system are developed on a training driving cycle that is built with elements generated from four standard driving cycles including Artemis Rural, RTS95, UDDS, and WLTP. As illustrated in Fig.4, the training cycle has four phases, where Phase 1 represents the low-speed region of the Artemis Rural cycle; Phase 2 involves the maximum acceleration in the RTS95 cycle; Phase 3 represents the medium-speed in UDDS Driving Cycle; Phase 4 involves the high-speed region in the WLTP Cycle. In each learning episode, the four phases will be reorganized randomly to provide the learning noise for robustness evaluation of the learning.

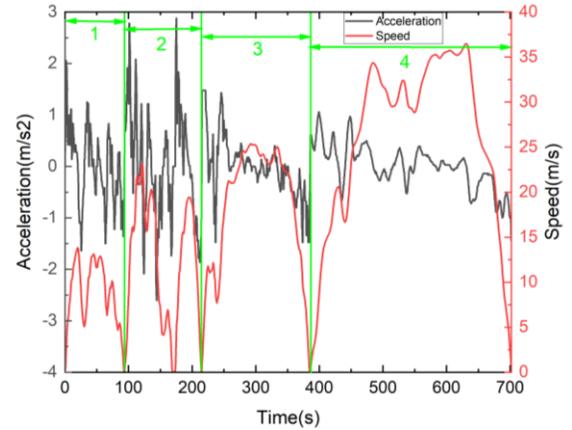

*Fig. 4. An example of the learning cycle*

*4.1 Learning performances of the multi-agent systems with different $R_{ind}$*

To determine the unified setting for the independence ratio in the multi-agent system, this paper investigated the performances of a PHEV controlled by the MADRL systems with the $R_{ind}$ value changing from 0 to 0.8 with 0.2 intervals. The simulation experiments are conducted under the learning cycles, and the learning curves of the two DDPG agents illustrating the evolution of agent rewards in 80 episodes are compared in Fig.5.

Most of the studied cases ($R_{ind} = 0, 0.2, \text{and } 0.6$) demonstrate the way of handshaking in the multi-agent system, in which, both agents have the same tendency of growing with their reward values increasing over time. Nevertheless, in the system with a $R_{\text{ind}}$ value of 0.4, the two agents are growing in exactly opposite directions. In the system with a $R_{\text{ind}}$ value of 0.8, although both agents have the same tendency, the reward values of the multi-agent system decrease over time. The study also suggested that the multi-agent system with a $R_{\text{ind}}$ value of 0.2 is the best since it can stably achieve the highest reward values for both agents with the fastest speed.

### 4.2 Comparison with the single-agent system

This section investigates the fuel economy and battery SoC performance of the multi-mode PHEVs controlled by the single-agent system and the multi-agent system with a $R_{\text{ind}}$ value of 0.2, respectively. Simulation experiments are conducted with different initial battery SoC values of 25%, 28%, and 30%, respectively. The results are summarized in Table II, in which the battery SoC error and energy-saving rate are calculated by

$$SOC_{error} = \frac{|SOC_{End} - SOC_{Initial}|}{SOC_{Initial}} * 100\%$$
$$Saving = \frac{|Fuel_{Multi-agent} - Fuel_{Single-agent}|}{Fuel_{Single-agent}} * 100\%$$
(21)

where $SOC_{initial}$ and $SOC_{End}$ are the battery SoC value at the beginning and the end of a driving cycle, respectively; $Fuel_{single-agent}$ and $Fuel_{Multi-agent}$ are the fuel consumption in L/100km for the single-agent system and the multi-agent system, respectively.

From the results illustrated in Table II, the proposed multi-agent system outperforms the single-agent system robustly. Compared to the single-agent system, the multi-agent system generates two control variables to allow MG1 and MG2 controlled independently. This mechanism enables more accurate control of battery SoC leading to less SoC error. The SoC error range of the multi-agent system is from 3.21% - 3.60% while the range of the single-agent system is 8.8% - 9.30%. In the three tests with different battery initial SoC, the multi-agent system saved 2.538%, 2.130%, and 2.554% fuels, respectively.

TABLE II
LEARNING PERFORMANCE OF THE SINGLE-AGENT AND MULTI-AGENT LEARNING

| Initial SoC | Method | End SoC | SOC Error (%) | Fuel/100km (L/100km) | Saving (%) |
|---|---|---|---|---|---|
| 25% | Single | 27.2% | 8.80 | 4.534 | - |
|  | Multi-agent | 24.1% | 3.60 | 4.419 | 2.538 |
| 28% | Single | 30.5% | 8.93 | 4.547 | - |
|  | Multi-agent | 27.1% | 3.21 | 4.450 | 2.130 |
| 30% | Single | 32.8% | 9.33 | 4.534 | - |
|  | Multi-agent | 28.6% | 4.67 | 4.418 | 2.554 |

### 5. CONCLUSION

This paper studied on learning-based energy management control of a multi-mode PHEV as a continuous MIMO control optimization problem, and a new multi-agent deep reinforcement learning system is proposed. The work has been down with two new contributions: 1) a hand-shaking strategy is proposed for the DRL agents by introducing an independence ratio, and 2) a parametric study is conducted to obtain the best setting for the MADRL framework. The conclusions drawn from the investigation are as follows:

1) The collaboration mode of the multi-agent system can be controlled by the proposed independence ratio, which needs to be fine-tuned otherwise the learning agents will grow in two opposite directions or go down together.
2) The study suggested that the unified setting for the independence ratio is 0.2 for control of the multi-mode PHEV concerning both learning speed and maximum reward achievable.
3) The proposed multi-agent system can reduce the battery SoC error by up to 59% while saving more

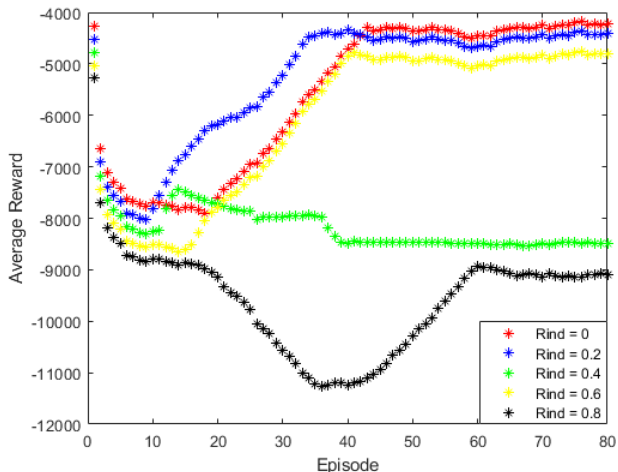
(a) Agent A

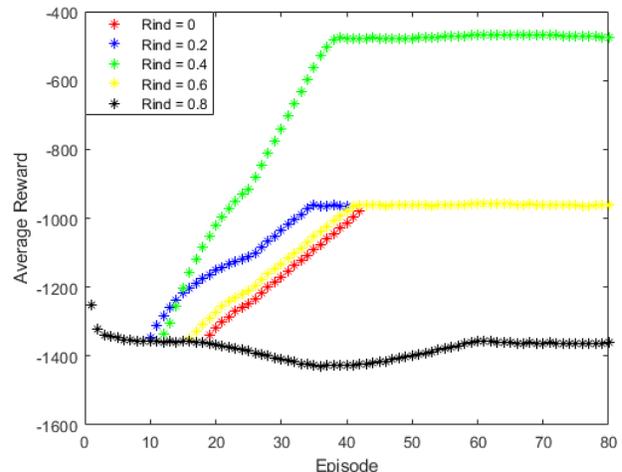
(b) Agent B

Fig.5. Learning performance of different ratios (SOC initial value is 0.28)

than 2.4% of fuel compared to the single-agent system.

**ACKNOWLEDGEMENT**

This work was supported by the National Natural Science Foundation of China (Grant No. 51976176) and the Open Fund Project from the State Key Laboratory of Clean Energy Utilization (No. ZJUCEU2021019).

**REFERENCE**

[1] Ganesh AH, Xu B. A review of reinforcement learning based energy management systems for electrified powertrains: Progress, challenge, and potential solution. Renew Sustain Energy Rev 2022;154:111833. https://doi.org/10.1016/j.rser.2021.111833.

[2] Hua M, Chen G, Zhang B, Huang Y. A hierarchical energy efficiency optimization control strategy for distributed drive electric vehicles. Proc Inst Mech Eng Part D J Automob Eng 2019;233:605–21. https://doi.org/10.1177/0954407017751788.

[3] Zhang F, Hu X, Langari R, Cao D. Energy management strategies of connected HEVs and PHEVs: Recent progress and outlook. Prog Energy Combust Sci 2019;73:235–56. https://doi.org/10.1016/j.pecs.2019.04.002.

[4] Chen G, Hua M, Zong C, Zhang B, Huang Y. Comprehensive chassis control strategy of FWIC-EV based on sliding mode control. IET Intell Transp Syst 2019;13:703–13. https://doi.org/10.1049/iet-its.2018.5089.

[5] Wirasingha SG, Emadi A. Classification and review of control strategies for plug-in hybrid electric vehicles. IEEE Trans Veh Technol 2011;60:111–22. https://doi.org/10.1109/TVT.2010.2090178.

[6] Odeim F, Roes J, Wülbeck L, Heinzel A. Power management optimization of fuel cell/battery hybrid vehicles with experimental validation. J Power Sources 2014;252:333–43. https://doi.org/10.1016/j.jpowsour.2013.12.012.

[7] Liu W, Member GS, Quijano K, Member GS, Crawford MM, Fellow L. YOLOv5-Tassel : Detecting Tassels in RGB UAV Imagery With Improved YOLOv5 Based on Transfer Learning. IEEE J Sel Top Appl Earth Obs Remote Sens 2022;15:8085–94. https://doi.org/10.1109/JSTARS.2022.3206399.

[8] Zhou Q, Zhao D, Shuai B, Li Y, Williams H, Xu H. Knowledge Implementation and Transfer With an Adaptive Learning Network for Real-Time Power Management of the Plug-in Hybrid Vehicle. IEEE Trans Neural Networks Learn Syst 2021;32:5298–308.

[9] Chen Z, Hu H, Wu Y, Zhang Y, Li G, Liu Y. Stochastic model predictive control for energy management of power-split plug-in hybrid electric vehicles based on reinforcement learning. Energy 2020;211:118931. https://doi.org/10.1016/j.energy.2020.118931.

[10] Lee H, Kang C, Park Y Il, Cha SW. A study on power management strategy of HEV using dynamic programming. World Electr Veh J 2016;8:274–80. https://doi.org/10.3390/wevj8010274.

[11] Chowdhury NR, Ofir R, Zargari N, Baimel D, Belikov J, Levron Y. Optimal Control of Lossy Energy Storage Systems with Nonlinear Efficiency Based on Dynamic Programming and Pontryagin's Minimum Principle. IEEE Trans Energy Convers 2021;36:524–33. https://doi.org/10.1109/TEC.2020.3004191.

[12] Sun C, Sun F, He H. Investigating adaptive-ECMS with velocity forecast ability for hybrid electric vehicles. Appl Energy 2017;185:1644–53. https://doi.org/10.1016/j.apenergy.2016.02.026.

[13] Qi X, Wu G, Boriboonsomsin K, Barth MJ. A Novel Blended Real-Time Energy Management Strategy for Plug-in Hybrid Electric Vehicle Commute Trips. IEEE Conf Intell Transp Syst Proceedings, ITSC 2015;2015-Octob:1002–7. https://doi.org/10.1109/ITSC.2015.167.

[14] Zhao S, Chen G, Hua M, Zong C. An identification algorithm of driver steering characteristics based on backpropagation neural network. Proc Inst Mech Eng Part D J Automob Eng 2019;233:2333–42. https://doi.org/10.1177/0954407019856153.

[15] Zhou Q, Li Y, Zhao D, Li J, Williams H, Xu H, et al. Transferable representation modelling for real-time energy management of the plug-in hybrid vehicle based on k-fold fuzzy learning and Gaussian process regression. Appl Energy 2022;305:117853. https://doi.org/10.1016/j.apenergy.2021.117853.

[16] Sun H, Fu Z, Tao F, Zhu L, Si P. Data-driven reinforcement-learning-based hierarchical energy management strategy for fuel cell/battery/ultracapacitor hybrid electric vehicles. J Power Sources 2020;455:227964. https://doi.org/10.1016/j.jpowsour.2020.227964.

[17] Xiong R, Cao J, Yu Q. Reinforcement learning-based real-time power management for hybrid energy storage system in the plug-in hybrid electric vehicle. Appl Energy 2018;211:538–48. https://doi.org/10.1016/j.apenergy.2017.11.072.

[18] Tang X, Chen J, Liu T, Qin Y, Cao D. Distributed Deep Reinforcement learning-based energy and emission management strategy for hybrid electric vehicles. IEEE Trans Veh Technol 2021;70:9922–34. https://doi.org/10.1504/IJVP.2022.119433.

[19] Liu W, Xia X, Xiong L, et al. Automated vehicle sideslip angle estimation considering signal measurement characteristic[J]. IEEE Sensors Journal, 2021, 21(19): 21675-21687.

[20] Abdullah HM, Gastli A, Ben-Brahim L. Reinforcement Learning Based EV Charging Management Systems-A Review. IEEE Access 2021;9:41506–31. https://doi.org/10.1109/ACCESS.2021.3064354.

[21] Zhou Q, Li J, Shuai B, Williams H, He Y, Li Z, et al. Multi-step reinforcement learning for model-free predictive energy management of an electrified off-highway vehicle. Appl Energy 2019;255. https://doi.org/10.1016/j.apenergy.2019.113755.

[22] Pei H, Hu X, Yang Y, Peng H, Hu L, Lin X. Designing Multi-Mode Power Split Hybrid Electric Vehicles Using the

Hierarchical Topological Graph Theory. IEEE Trans Veh Technol 2020;69:7159–71. https://doi.org/10.1109/TVT.2020.2993019.
[23] Xu R, Li J, Dong X, et al. Bridging the Domain Gap for Multi-Agent Perception[J]. arXiv preprint arXiv:2210.08451, 2022.
[24] Chen W, Xu R, Xiang H, et al. Model-Agnostic Multi-Agent Perception Framework[J]. arXiv preprint arXiv:2203.13168, 2022.
[25] Schmidt L M, Brosig J, Plinge A, et al. An Introduction to Multi-Agent Reinforcement Learning and Review of its Application to Autonomous Mobility[J]. arXiv preprint arXiv:2203.07676, 2022.
[26] Lowe R, Wu Y I, Tamar A, et al. Multi-agent actor-critic for mixed cooperative-competitive environments[J]. Advances in neural information processing systems, 2017, 30.